\documentclass[10pt,twocolumn,letterpaper]{article}

\usepackage{iccv}
\usepackage{times}
\usepackage{epsfig}
\usepackage{graphicx}
\usepackage{amsmath}
\usepackage{amssymb}
\usepackage{ragged2e}
\usepackage{multicol}
\usepackage{multirow}
\usepackage[misc]{ifsym}
\usepackage{graphicx} 
\usepackage{float} 
\usepackage{subfigure} 
\usepackage{tabularx}
\usepackage{booktabs}
\usepackage{wasysym}
\usepackage{tensor}
\usepackage[para,online,flushleft]{threeparttable}
\newcommand{\Rho}{\mathrm{P}}
\usepackage[symbol]{footmisc}
\usepackage[accsupp]{axessibility}

\usepackage{hyperref}
\hypersetup{breaklinks=true,colorlinks}
\iccvfinalcopy 

\def\iccvPaperID{2217} 

\ificcvfinal\fi

\begin{document}

\title{Learning Canonical View Representation for 3D Shape Recognition with Arbitrary Views}

\author{Xin Wei\textsuperscript{1*}, 
Yifei Gong\textsuperscript{2*}, 
Fudong Wang\textsuperscript{2}, 
Xing Sun\textsuperscript{2{~\small\Letter~}}, 
Jian Sun\textsuperscript{1{~\small\Letter}}\\
\textsuperscript{1}Xi'an Jiaotong University, \textsuperscript{2} Tencent Youtu Lab\\
{\tt\small wxmath@stu.xjtu.edu.cn, \{yifeigong,winfredsun\}@tencent.com} \\ {\tt\small fudong-wang@whu.edu.cn, jiansun@xjtu.edu.cn}}
\maketitle
\renewcommand{\thefootnote}{\fnsymbol{footnote}}
\footnotetext[1]{Equal contribution.}
\ificcvfinal\thispagestyle{empty}\fi

\begin{abstract}
    In this paper, we focus on recognizing 3D shapes from arbitrary views, i.e., arbitrary numbers and positions of viewpoints. It is a challenging and realistic setting for view-based 3D shape recognition. We propose a  
    canonical view representation to tackle this challenge. We first transform the original features of arbitrary views to a fixed number of view features, dubbed canonical view representation, by aligning the arbitrary view features to a set of learnable reference view features using optimal transport. In this way, each 3D shape with arbitrary views is represented by a fixed number of canonical view features, which are further aggregated to generate a rich and robust 3D shape representation for shape recognition. We also propose a 
    canonical view feature separation constraint to enforce that the view features in canonical view representation can be embedded into scattered points in a Euclidean space. Experiments on the ModelNet40, ScanObjectNN, and RGBD datasets show that our method achieves competitive results under the fixed viewpoint settings, and significantly outperforms the applicable methods under the arbitrary view setting. 

\end{abstract}
\section{Introduction}
Understanding the 3D world is a fundamental problem in computer vision. One of its central challenges is how to represent and recognize objects in the 3D space. Recently, many view-based methods~\cite{chen2020learning,emv,hypergraph,gvcnn,n_gram,deepccfv,rotationnet,mvcnn,mvcnn_new,dscnn,view-gcn,relationnet,mhbn} were proposed to recognize 3D shape with multi-view 2D images based on the aggregation of features learned by deep neural networks. Leveraging advances in 2D image descriptors (\eg~\cite{resnet}) and massive image databases~\cite{imagenet}, they are among the state-of-the-art methods for 3D shape recognition.  

\begin{figure} 
\label{fig:first} 
\includegraphics[width=0.48\textwidth]{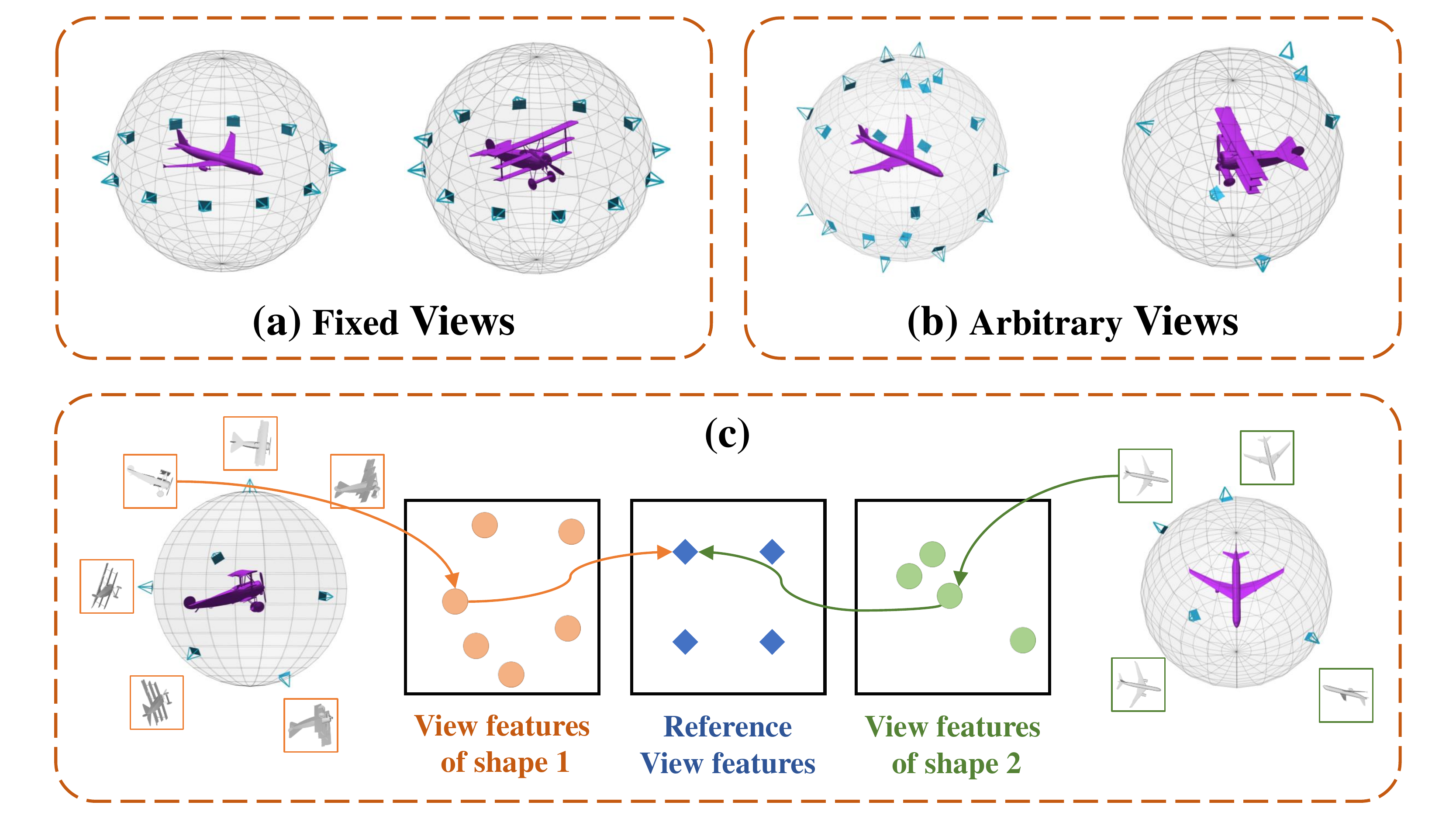} 
\caption{This paper addresses 3D shape recognition with arbitrary views as shown in (b), which is more challenging and realistic than the fixed-viewpoint setting in (a). As shown in (c), given an arbitrary number of unaligned view images, our method learns canonical view features of a 3D shape aligned to a fixed number of learnable reference view features using optimal transport.}

\label{fig:fig1} 
\end{figure}

However, most of these methods~\cite{chen2020learning,emv,hypergraph,gvcnn,n_gram,rotationnet,mvcnn,mvcnn_new,dscnn,view-gcn,relationnet,mhbn} focus on settings with a pre-defined camera setup where the same set of viewpoints are used for every object, \eg, Fig.~\ref{fig:first}(a). 
In practical applications, 3D objects are often observed from arbitrary views without knowing their precise camera positions. 
In this work, we aim to tackle 3D shape recognition with arbitrary views. The setting can be defined as follows.  
(i) Views are taken from arbitrary viewpoints for each object. (ii) Objects have varying numbers of observation views, \eg, Fig.~\ref{fig:first}(b).

Compared with the fixed-viewpoint setup, 3D shape recognition faces new challenges brought by the unaligned inputs from arbitrary views. 
It is difficult to robustly aggregate features of structurally unaligned views. Moreover, representations learned from a typical neural network are also mutually-unaligned in the feature space,  
where feature aggregation could result in a loss of discriminability. 

To tackle these challenges, an intuitive motivation is to recover the inherent alignment for the arbitrary views. Specifically, if we find a link between the unaligned features from arbitrary views and a set of virtual reference views for observing an object, we can transform the features into aligned representations for the subsequent aggregation. 



Driven by this motivation, we design a novel canonical view representation for 3D shape recognition with arbitrary views. Specifically, the input arbitrary views of each 3D shape are first processed by an image-level feature encoder consisting of a CNN and a Transformer encoder~\cite{attention}. Then these features of arbitrary views are transformed into canonical view features aligned to a fixed number of learned reference view features. The transformation mapping is derived by the optimal transport~\cite{Sinkhorn2013,ROT1,OT_old}. To ensure that the canonical view features are distinct, we require that the canonical view features can be embedded into a Euclidean space (\eg, $\mathbb{R}^3$) with mutually distant coordinates. In this way, each 3D shape is represented by a fixed number of features over the reference views in the feature space, resulting in the canonical representation of each 3D shape.  The aligned canonical view features added with spatial embeddings are further encoded and aggregated to generate a discriminative global representation of the 3D shape. 

Our main contributions can be summarized as follows.
We tackle the challenge of 3D object recognition with arbitrary views by introducing a novel canonical view representation, which recovers the inherent mutual-alignment features among arbitrary views and produces a rich representation of the 3D shape.
We further propose a canonical view feature separation loss to ensure feature separability, which improves the discriminability and robustness of the final representation. We conduct experiments on CAD, scanned model and real-world image datasets including ModelNet40~\cite{modelnet}, ScanObjectNN~\cite{scanobjectnn} and RGBD~\cite{rgbd} dataset. The results show that our approach significantly outperforms the state-of-the-art methods under the challenging setting of 3D shape recognition with arbitrary views.
\begin{figure*} 
\centering 
\includegraphics[width=1.0\textwidth]{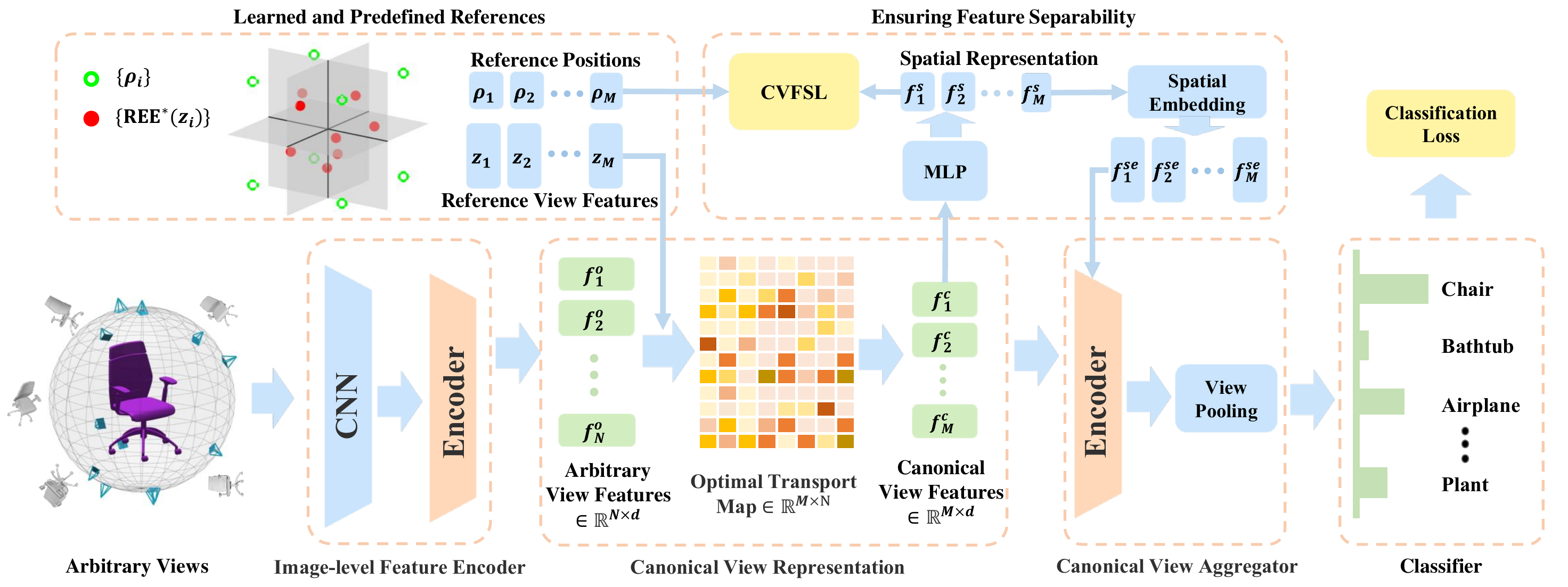} 
\caption{Overview of our approach. The network consists of three components, i.e., Image-level Feature Encoder (ILFE), Canonical View Representation (CVR), and Canonical View Aggregator (CVA). Images from $N$ arbitrary views are first encoded by the ILFE, then the original unaligned features $F^o=\{f^o_i\}_{i=1}^N$ are transformed into a fixed number $M$ of canonical view features $F^c=\{f^c_i\}_{i=1}^M$ aligned to the  
learned reference view features $Z=\{z_i\}_{i=1}^M$. 
Optimal transport is performed between $F^o$ and $Z$ to obtain the canonical view features $F^c$, while a novel Canonical View Feature Separation Loss (CVFSL) ensures canonical view features $F^c$ to be distinct and separable. 
The CVA with spatial embeddings further explores the inter-viewpoint relationship and aggregates the canonical view features. *Robust Euclidean Embedding~(REE) is used to visualize $Z$ in an example 3D space.}

%
\label{fig:main} 
\end{figure*}
\section{Related Work}

\subsection{3D shape recognition with multi-view images}
View-based methods in 3D shape recognition have proved to be effective while only requiring 2D input images observed from different viewpoints. The key challenge of the view-based methods is how to effectively aggregate the features of multiples views to generate the shape descriptor. 

MVCNN~\cite{mvcnn} is a framework that aggregates multi-view features with max-pooling, achieving superior performance against methods directly working on 3D inputs. Multi-view feature aggregation is further explored in GVCNN~\cite{gvcnn} where the view features are grouped to obtain more informative representation. Similarly, view-GCN~\cite{view-gcn} uses Graph Convolutional Neural Network to model the relations among different viewpoints to hierarchically aggregate features of multiple views.  
RotationNet~\cite{rotationnet} attempts to tackle the challenge of perturbed objects by predicting the object pose to represent the 3D shape in its aligned form. EMV\cite{emv} also tries to solve this problem with group convolution over discrete rotation groups. While achieving impressive performance, these approaches assume having a pre-defined set of viewpoints for each object. This makes them unfitted for the more practical setting where viewpoint positions are arbitrary and different for every object.

To the best of our knowledge, there are few works that go beyond the fixed viewpoints setup. DeepCCFV~\cite{deepccfv} tries to simulate a constraint-free camera setup in the testing phase and improve the generalization performance. However, it still assumes a pre-defined camera setup for the training data and the retrieval gallery, and the queries are sampled from the pre-defined viewpoints. 
OVCNet~\cite{liu2020recognizing} attempts to tackle the task of shape recognition from any view, but mainly targets at the single-view scenario and relies on a challenging task of 3D reconstruction from a single image, while our method focuses on effectively aggregating multi-view images from arbitrary viewpoints.

Compared with the above-mentioned methods, our proposed method is also view-based, but we flexibly relax the fixed viewpoints setup to the arbitrary viewpoints setup. Our method achieves the state-of-the-art results in this challenging setting by employing a canonical view representation that  
aligns the image-level features of arbitrary views to a set of reference view features. 
\subsection{Transformer networks}
Transformers~\cite{language,bert,attention} are initially introduced as an encoder-decoder architecture for machine translation, where the self-attention mechanism is incorporated to model the relationship among a set of inputs. To model the positional information of the sequential inputs, positional encodings are added to the input embedding. They are widely adopted in NLP for their scalability and good generalization performance.

Transformer networks are also proved to be effective for computer vision tasks~\cite{dert,ipt,vit,pct,swintransformer,pyramidtransformer,point_transformer}. DETR~\cite{dert} is an object detection method based on Transformer, which encodes the image features and decodes the objects in parallel. VIT~\cite{vit} demonstrated the feasibility of using Transformer as the backbone for image  classification, and outperforms the popular CNNs. 

In this work, we first utilize the Transformer encoder~\cite{attention} as an effective way to explore the relationship among features of arbitrary views. After we  
transform the features into the canonical view representation, we use another Transformer encoder~\cite{attention} to process the aligned canonical view features added with spatial embeddings, resulting in the final representation of the 3D shape.



\section{Canonical View Representation} \label{sec:CR}

We first introduce our proposed canonical view representation for 3D shape recognition with arbitrary views, taken as the basis of our 3D shape recognition network presented in Sect.~\ref{sec:net}. The major objective of this representation is to transform a set of arbitrary view features of a 3D shape to be a fixed number of view features, by learning and aligning to the same number of reference view features in the feature space. The optimally transformed features are dubbed \textit{\textbf{canonical view representation}} of a 3D shape.

\begin{figure*} 
\centering 
\includegraphics[width=0.98\textwidth]{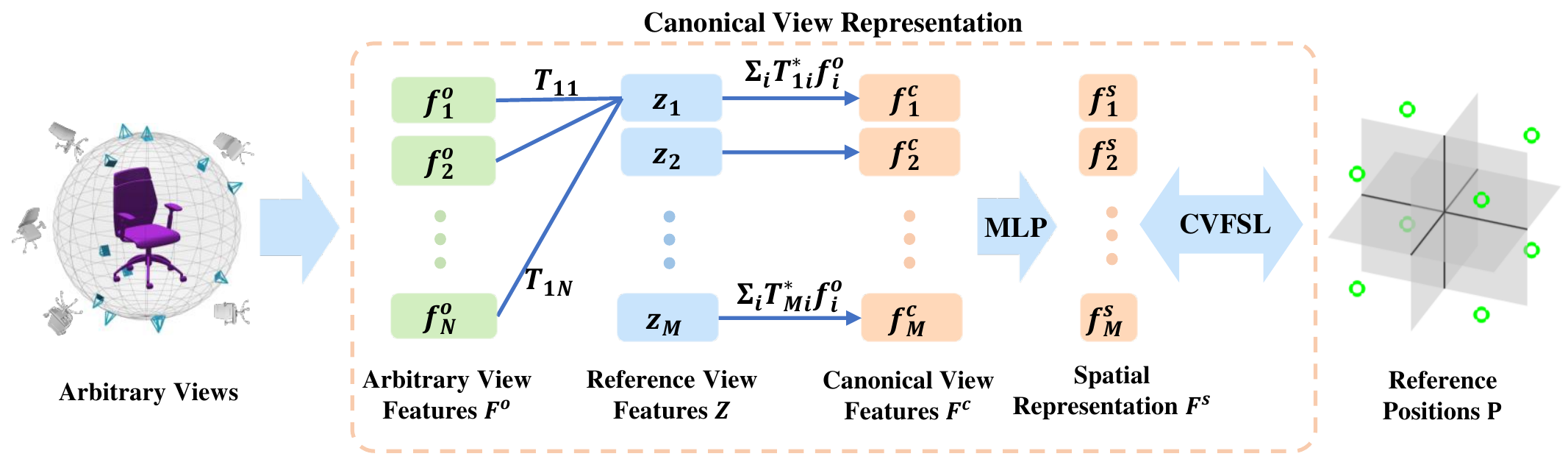} 
\caption{Illustration of the canonical view representation for 3D shape with arbitrary views. Given the unaligned features $F^o\triangleq\{f^o_i\}_{i=1}^N$ and reference view features $Z\triangleq\{z_j\}_{j=1}^M$, the transformation map $\mathbf{T}^*\triangleq\{T^*_{ji}\}$ is calculated with optimal transport in Eq.~\eqref{eq:ROT}. The transformed feature $F^c\triangleq\{f^c_i\}_{i=1}^M$ are the aligned canonical view features. The Canonical View Feature Separation Loss (CVFSL) ensures that $F^c$ are aware of the reference positions $\{\Rho=\rho_i\}_{i=1}^M$.}
\label{fig:optimal_transport} 
\end{figure*}

Suppose that we have extracted features from each view of 3D shape by the Image-level Feature Encoder (in Sect.~\ref{sec:ILF-encoder}). We next present how to model a set of reference view features in the feature space and transform the arbitrary view features to a canonical view representation based on optimal transport, as shown in the  Fig.~\ref{fig:optimal_transport}. 
In order to increase the discriminative ability of the canonical view representation, we also propose a constraint to ensure that the canonical view representations are separable in the feature space. Since the involved computations are differentiable, the computations for the canonical view representation will be taken as network modules in our 3D shape recognition network introduced in Sect.~\ref{sec:net}, and the reference view features and sub-nets in canonical view representation can be learned by network training. 




\subsection{Formulation}
Given varying $N$ arbitrary views of a 3D shape, we first extract their original features $F^{o}\triangleq\{f^o_i\}_{i=1}^N \in \mathbb{R}^{N\times d}$ by an image-level feature encoder (in Sect.~\ref{sec:ILF-encoder}). Then, to obtain a fixed number ($M$) of view features from the features $F^{o}$ of  $N$ arbitrary views, we propose to find a feature transform $\mathcal{T}: \mathbb{R}^{N\times d} \to \mathbb{R}^{M\times d}$ such that $\mathcal{T}(F^o)\in\mathbb{R}^{M\times d}$. We assume that $\mathcal{T}$ is linear which is reasonable in high-dimensional feature space. Now we have
\begin{equation}
    \mathcal{T}(F^o)\triangleq \mathbf{T}F^o, \quad f^t_j\triangleq\sum_{i}\mathbf{T}_{ji}f^o_i, \forall j=1,...,M,
\end{equation}
where $\mathbf{T}\in\mathbb{R}^{M\times N}$ is a linear transform map implementing $\mathcal{T}$ and $F^{t}\triangleq\{f^t_j\}_{j=1}^M \in \mathbb{R}^{M\times d}$ are the transformed features taken as the candidate canonical view representations. We hope to find an optimal transform map $\mathbf{T}^*$ to construct $F^{t}$, which is detailed in the followings.

\textbf{Reference view representation}. We further specify the transform $\mathbf{T}$ as the mapping from $N$ arbitrary view features to a fixed number ($M$) learnable reference view features $Z\triangleq\{z_j\}_{j=1}^M$, which can be seen as virtual reference views shared by all different 3D shapes. We define a similarity function $S(f^t_i, z_j)$ to measure similarity between $f^t_i$ and $z_j$ for $i\in[1, N], j \in [1, M]$, and solve the following optimization problem to find an optimal transform map $\mathbf{T}^*$:
\begin{equation}
    \mathbf{T}^*\triangleq\mathop{\text{argmax}}\limits_{\mathbf{T}}\sum_{j}S(f^t_j, z_j)=\sum_{j}S(\sum_{i}\mathbf{T}_{ji}f^o_i, z_j).\label{eq:similar} 
\end{equation}
In this paper, a simple yet efficient definition of $S(\cdot,\cdot)$ is adopted as the linear inner product
$   S(f^t_i, z_j)\triangleq f^t_i\cdot z_j.$

\textbf{Optimal transport solver}. Due to the linearity of $S$, the optimization problem in Eq.~\eqref{eq:similar} can be rewritten as
\begin{equation}
\mathbf{T}^*\triangleq\mathop{\text{argmin}}\limits_{\mathbf{T}}\sum_{ij}-\mathbf{T}_{ji}S(f^o_i, z_j),\label{eq:LP}
\end{equation}
which can be solved by many linear programming algorithms~\cite{LP2,LP_inter}. However, to guarantee the regularization for $\mathbf{T}$ and differentiability for the training procedure, we regularize $\mathbf{T}$ to be a doubly-stochastic matrix~\cite{doublely1959,sinkhorn1967} and add an entropy-based regularization term to Eq.~\eqref{eq:LP}:
\begin{equation}
    \mathbf{T}^*\triangleq\mathop{\text{argmin}}\limits_{T}\sum_{ij}-\mathbf{T}_{ji}S(f^o_i, z_j)+\epsilon\sum_{ij}\mathbf{T}_{ji}\text{ln}(\mathbf{T}_{ji}),\label{eq:ROT}
\end{equation}
where $\epsilon\ge 0$ is a balance weight. Moreover, Eq.~\eqref{eq:ROT} is a well-known regularized optimal transport problem~\cite{bregmen,ROT1,OTK}, that can be solved differentiably with the Sinkhorn algorithm~\cite{Sinkhorn2013}.

\textbf{Canonical view representation}. Once the optimal $\mathbf{T}^*$ is solved, we get the canonical view features as $F^c\triangleq\{f^c_j\}_{j=1}^M$, where $f^c_j=\sum_{i}\mathbf{T}^*_{ji}f^o_i$. Thus the canonical view representation of a 3D shape with arbitrary views are the optimally transformed features under the alignment constraint w.r.t. the reference view features.



\subsection{Canonical View Feature Separability} 
The canonical view representation obtained above for each 3D shape is length-and-order fixed benefiting from the reference view representation, 
but the resulting features might suffer from homogenization 
without proper constraint. Thus we propose the \textbf{C}anonical \textbf{V}iew \textbf{F}eature \textbf{S}eparation \textbf{L}oss (CVFSL) to instill separability among these features. More precisely, we require that the canonical view representation $F^c$ of a 3D shape can be embedded into a spatial representation $F^s\in\mathbb{R}^{M\times k}$ such that $F^s$ are scattered in the $k$-dimensional Euclidean space. 

To achieve this goal, we utilize a two-layer MLP network $\Phi(\cdot)$ with a hidden dimension of 64 to extract the spatial representation $F^s\in \mathbb{R}^{M\times k}$ from $F^c \in \mathbb{R}^{M \times d}$, such that $
    F^s=\Phi(F^c).
$
To make the spatial representation $F^s$ scatter uniformly in the $\mathbb{R}^k$ space, we enforce the constraint that
\begin{equation}
    \text{L}_{sep}\triangleq \sum_{j=1}^M||f^{s'}_j-\frac{\rho_j}{||\rho_j||}||_2^2,\label{eq:spatial-loss},
\end{equation}
where $f^{s'}$ is the $l_2$-normalization of $f^{s}$, and reference positions $\Rho\triangleq \{{\rho}_j\}_{j=1}^M\in\{1,-1\}^k, M=2^k$. When training our network (in Sect.~\ref{sec:net}) using this loss as one term,  
it enforces that the canonical view representation of each 3D shape are separable and discriminative. The effectiveness of this design is validated in Sect.~\ref{sec:ablation}.

\section{Network Architecture} \label{sec:net}
As shown in Fig.~\ref{fig:main}, our network for 3D shape recognition consists of three modules: the Image-level Feature Encoder (ILFE), the Canonical View Representation (CVR), and the Canonical View Aggregator (CVA). The Image-level Feature Encoder is composed of the CNN backbone and a Transformer encoder~\cite{attention}. Given \textit{N} arbitrary views $\{I_i\}^N_{i=1}$, the CNN backbone processes each view individually, and the Transformer encoder further processes the whole set of views to output a richer feature for each view, denoted by $F^{o}\triangleq\{f^o_i\}_{i=1}^N$. The Canonical View Representation module aims to align the features in $F^o$ of arbitrary views to the reference view features $Z\triangleq\{z_j\}_{j=1}^M$ that are also learned during training. 
We compute a linear transformation map $\mathbf{T}^*\in\mathbb{R}^{M \times N}$ using optimal transport. It can optimally transform $F^o$ to a fixed sized canonical view representation $F^{c}\triangleq\{f^c_i\}_{i=1}^M$ based on reference view features $Z$. 
$F^{c}$ are then processed by the Canonical View Aggregator to derive a global feature for the 3D shape. In the CVA, a transformer encoder explores the relationship among the view features of the canonical view representation with spatial embedding, followed by a Global Average Pooling (GAP) layer to obtain the global feature for the 3D shape.  We next introduce these network modules.


\subsection{Image-level Feature Encoder}\label{sec:ILF-encoder}
As shown in Fig.~\ref{fig:main}, the image-level feature encoder consists of the CNN backbone and the transformer encoder. Given \textit{N} views $\{I_i\}^N_{i=1}$, the CNN backbone processes the images individually and produces the view features $F^v\triangleq\{f^v_i\}^N_{i=1} \in \mathbb{R}^{N\times d}$. The features $F^v$ are then processed by a transformer encoder~\cite{attention}, where multi-head self-attention and the Feed-Forward Network (FFN) are utilized to extract the information among arbitrary views. In the self-attention layer, the queries, keys and values are obtained by linearly projecting the view features. Namely, the query $Q$, key $K$ and value $V$ are denoted as
 \begin{equation}
     Q \triangleq F^vW^Q, \quad K \triangleq F^vW^K, \quad V \triangleq F^vW^V,
 \end{equation}
where $W^Q \in \mathbb{R}^{d \times d}$, $W^K \in \mathbb{R}^{d \times d}$, and $W^V \in \mathbb{R}^{d \times d}$ are learnable linear weights. We utilize the Scaled Dot-Product Attention~\cite{attention} defined as
\begin{equation}
    \text{Attention}(Q,K,V) = \text{softmax}(\frac{QK^T}{\sqrt{d_k}})V
\end{equation}
Then the Multi-Head Attention (MHA) is calculated as
\begin{align}\label{eq:MHA} 
\begin{split}
    \text{MHA}(Q,K,V) &= \text{Concat}(head_1, ..., head_h)W^O \\
    \text{where \:}head_h &= \text{Attention}(Q, K, V) 
\end{split}
\end{align}
Here $W^o\in \mathbb{R}^{hd \times d}$ reduces the dimension of the concatenated attention heads. The relationship among the arbitrary views are explored. 
The results are fed into an FFN~\cite{attention}, from which we obtain the image-level features denoted as $F^{o}\triangleq\{f^o_i\}_{i=1}^N$ of the input arbitrary views. FFN$(\cdot)$~\cite{attention} is a simple neural network using a two-layer MLP following the standard Transformer architecture. There are also residual connections and layer normalization~\cite{layernorm} after every block.

\subsection{Canonical View Representation}\label{sec:CR-decoder}
As demonstrated in Sect.~\ref{sec:CR}, the Canonical View Representation (CVR) module consists of three main operations, including (i)~learning the reference view features $Z$, (ii)~transforming the image-level features $F^o$ into the canonical view features $F^c$ with optimal transport,  
(iii)~ensuring separability of the canonical view representation with CVFSL. The illustration of the process is shown in Fig.~\ref{fig:optimal_transport}.

\noindent{\textbf{Update of $Z$}}. We first randomly initialize it as $Z^0\in \mathbb{R}^{M\times d}$. Then, with the forwarded features $F^o\in \mathbb{R}^{N\times d}$, we construct the objective function in Eq.~\eqref{eq:ROT} that is solved differentiably with the Sinkhorn algorithm. In this way, the gradient of both $Z$ and $F^o$ can be calculated so as to update $Z$ during training.

\noindent{\textbf{Transformation of $F^o$}}. Given $F^{o}$, we calculate the canonical view feature as the linearly-transformed features of $F^o$ with the optimal transport map:
\begin{equation}
    F^c = \mathbf{T}^*F^o,
\end{equation}
where $\mathbf{T}^*$ is the solution of Eq.~\eqref{eq:ROT}.


{\textbf{Separability constraint on $F^c$}}.
With the canonical view features $F^c$, a two-layer MLP with hidden dimension of 64 is used to extract the spatial representation $F^{s}\in\mathbb{R}^{M \times k}$, by $F^s=\text{MLP}(F^c)$. Then, we construct the canonical view feature separation constraint in Eq.~\eqref{eq:spatial-loss} to enforce that $F^s$, inferred from the canonical view representation $F^c$, scatters uniformly in the $\mathbb{R}^{M \times k}$ space.

\subsection{Canonical View Aggregator} \label{sec:CF-aggregator}
The Canonical View Aggregator (CVA) further processes the canonical view features $F^c$ along with the spatial representation $F^s$, and produces a global representation of the 3D shape. Given the canonical view representation $F^c\in\mathbb{R}^{M \times d}$, we explore the relationship between view features in the canonical view representation and aggregate them into a global feature $F^g$ for the 3D shape.  

\noindent{\textbf{Transformer encoder with spatial embedding.}} Given the spatial representation $F^s\in\mathbb{R}^{M \times k}$ calculated in the CVR, we obtain the spatial embedding $F^{se}\in\mathbb{R}^{M \times d}$ by $F^{se} = \Psi(F^s)$, where $\Psi(\cdot)$ is designed as a two-layer MLP network with 64 hidden units and a LeakyReLU layer. Thus we calculate the query $Q$, key $K$, and value $V$ by
\begin{equation}\label{eq:qkv_cra}
    \begin{aligned}
    Q &\triangleq (F^c+F^{se})W^Q, \\
    K &\triangleq (F^c+F^{se})W^K, \\
    V &\triangleq F^cW^V.
    \end{aligned}
\end{equation}
We then calculate the multi-head attention as in Eq.~\eqref{eq:MHA}, after which the outputs are fed into a Feed Forward Network, resulting in the same number of features $F^{ce}\in\mathbb{R}^{M \times d}$. 

\noindent{\textbf{Global representation of the 3D shape.}} We obtain the global representation $f^g\in\mathbb{R}^{1 \times d}$ of the 3D shape by performing Global Average Pooling (GAP) on the outputs of the Transformer encoder $F^{ce}$, by $f^g=\text{GAP}(F^{ce})$.


\subsection{Classifier}
We construct the classification module by a two-layer MLP network with hidden dimension of $\frac{d}{2}$. The output of the MLP is then fed into a softmax layer and the resulting logits represent the probability of each class.

\subsection{Network Training}
\noindent{\textbf{Training Loss.}} The training loss of our network consists of the classification loss $\text{L}_{cls}$ and the Canonical View Feature Seperation Loss $\text{L}_{sep}$. The overall loss is defined as
\begin{equation}\label{eq:overall_loss}
    \begin{aligned}
        \text{L} &\triangleq \text{L}_{cls} + \text{L}_{sep} \\
        &= -\sum_{c=1}^Cy_c\text{log}\:p_c + \lambda(\sum_{j=1}^M||f^{s'}_j-\frac{\rho_j}{||\rho_j||}||_2^2),
    \end{aligned}
\end{equation}
where $y_c$ and $p_c$ are the true probability and predicted probability of class $c$, while $f_j^{s'}$ and $\rho_j$ are  defined in Eq.~\eqref{eq:spatial-loss}. 

\noindent{\textbf{Hyper-parameters and backbone.}} We adopt the ResNet-18~\cite{resnet} pretrained on ImageNet\cite{imagenet} as our CNN backbone network in Image-level Feature Encoder and set the feature dimension to $d=512$. $M=2^k$ is the number of canonical views, which affects how the Canonical View Representation module processes inputs. We use $M=8, k=3$ for experiments on ModelNet40~\cite{modelnet} and ScanobjectNN~\cite{scanobjectnn}, while $M=4,k=2$ for RGBD~\cite{rgbd}. Further discussion on the effect of $M$ is in Sect.~\ref{sec:ablation}. $\lambda$ is the weighting factor for the canonical view feature separation loss as in Eq.~\eqref{eq:overall_loss}, which we set $\lambda=0.1$ for the experiments. $\epsilon$ is the balance weight defined in \eqref{eq:ROT}, empirically set to 0.05.

\noindent{\textbf{Training Details.}} For all the experiments, we train our network for 60 epochs, with a batch-size of 20 on a NVIDIA V100 GPU. For the aligned and rotated setting, each batch contains 20 shapes with 400 multi-view images. For the arbitrary-view setting, the number of views for each shape varies. Variable-length view features from the CNN backbone network are zero-padded to the max number of views (20) and batched together. We use SGD with momentum as the optimizer. The initial learning rate, weight decay, momentum are $10^{-3}$, $10^{-3}$, 0.9 respectively. The learning rate follows the warm-up strategy~\cite{warmup} in the first epoch, and it linearly increases from 0 to $10^{-3}$. Then it is reduced to $10^{-5}$ following a cosine quarter-cycle. Our code will be available on \url{http://github.com/weixmath/CVR}.
\begin{table*}[]
        \caption{Shape classification accuracy (in $\%$) on ModelNet40. `NA' / `-': method is not applicable or result was not reported.}
        \begin{center}
        \small
        \setlength{\tabcolsep}{2mm}{
        \begin{tabular}{lcccccc}
        \toprule
                \hline
                \multirow{2}{*}{Method} & \multicolumn{2}{c}{Aligned}  & \multicolumn{2}{c}{Rotated}          & \multicolumn{2}{c}{Arbitrary Views}          \\ \cline{2-7} 
                & Per Class Acc.   & {Per Ins. Acc.} & Per Class Acc.   & Per Ins. Acc.     & Per Class Acc.   & Per Ins. Acc.    \\ \hline
                MVCNN-M                & 94.30\%          & 96.35\%          & 87.95\%          & 88.17\%          & 78.86\%          & 83.20\%          \\ 
                GVCNN-M                & 94.46\%          & 96.07\%          & 89.69\%          & 88.10\%          &     80.98\%            & 83.57\%                  \\ 
                RotationNet~\cite{rotationnet}             & -          & 97.37\%          & 84.74\%          & 85.29\%          & NA               & NA                \\ 
                View-GCN~\cite{view-gcn}                & \textbf{96.50\%} & \textbf{97.60\%}          & 85.90\%          & 88.25\%          & NA                & NA               \\ \hline
                Ours                    & 95.77\%          & 97.16\% & \textbf{91.12\%} & \textbf{92.22\%} & \textbf{84.01\%} & \textbf{86.91\%} \\ \hline
                \bottomrule
        \end{tabular}}
\end{center}
\vspace{-0.35cm}
\label{table:modelnet}
\end{table*}

\begin{table*}[]
        \caption{Shape classification accuracy (in $\%$) on ScanObjectNN. `NA' represents that the method is not applicable to this setting.}
        \begin{center}
        \small
        \setlength{\tabcolsep}{2mm}{
        \begin{tabular}{lcccccc}
                        \toprule
                        \hline
                        \multirow{2}{*}{Method} & \multicolumn{2}{c}{Aligned}                           & \multicolumn{2}{c}{Rotated}          & \multicolumn{2}{c}{Arbitrary Views}          \\ \cline{2-7} 
                        & Per Class Acc.   & {Per Ins. Acc.} & Per Class Acc.   & Per Ins. Acc.     & Per Class Acc.   & Per Ins. Acc.    \\ \hline
                        MVCNN-M                & 85.71\%          & 87.82\%                            & 78.21\%          & 80.62\%          & 58.58\%          & 63.29\%          \\ 
                        GVCNN-M                & 86.64\%          & 88.68\%                            & 82.86\%          & 83.70\%          &  58.84\%              & 65.35\%               \\ 
                        RotationNet~\cite{rotationnet}             & 84.88\%          & 86.90\%                            & 74.68\%          & 76.16\%          & NA                & NA                \\ 
                        View-GCN~\cite{view-gcn}                & \textbf{88.67\%} & 90.39\%                            & 81.99\%          & 83.50\%          & NA                & NA               \\ \hline
                        Ours                    & 88.39\%          & \textbf{90.74\%}                   & \textbf{84.70\%} & \textbf{85.59\%} & \textbf{68.07\%} & \textbf{71.36\%} \\ \hline
                        \bottomrule
                \end{tabular}}
        \end{center}
        \vspace{-0.3cm}
\label{table:scanobject}
\end{table*}

\begin{figure}  
\includegraphics[width=0.47\textwidth]{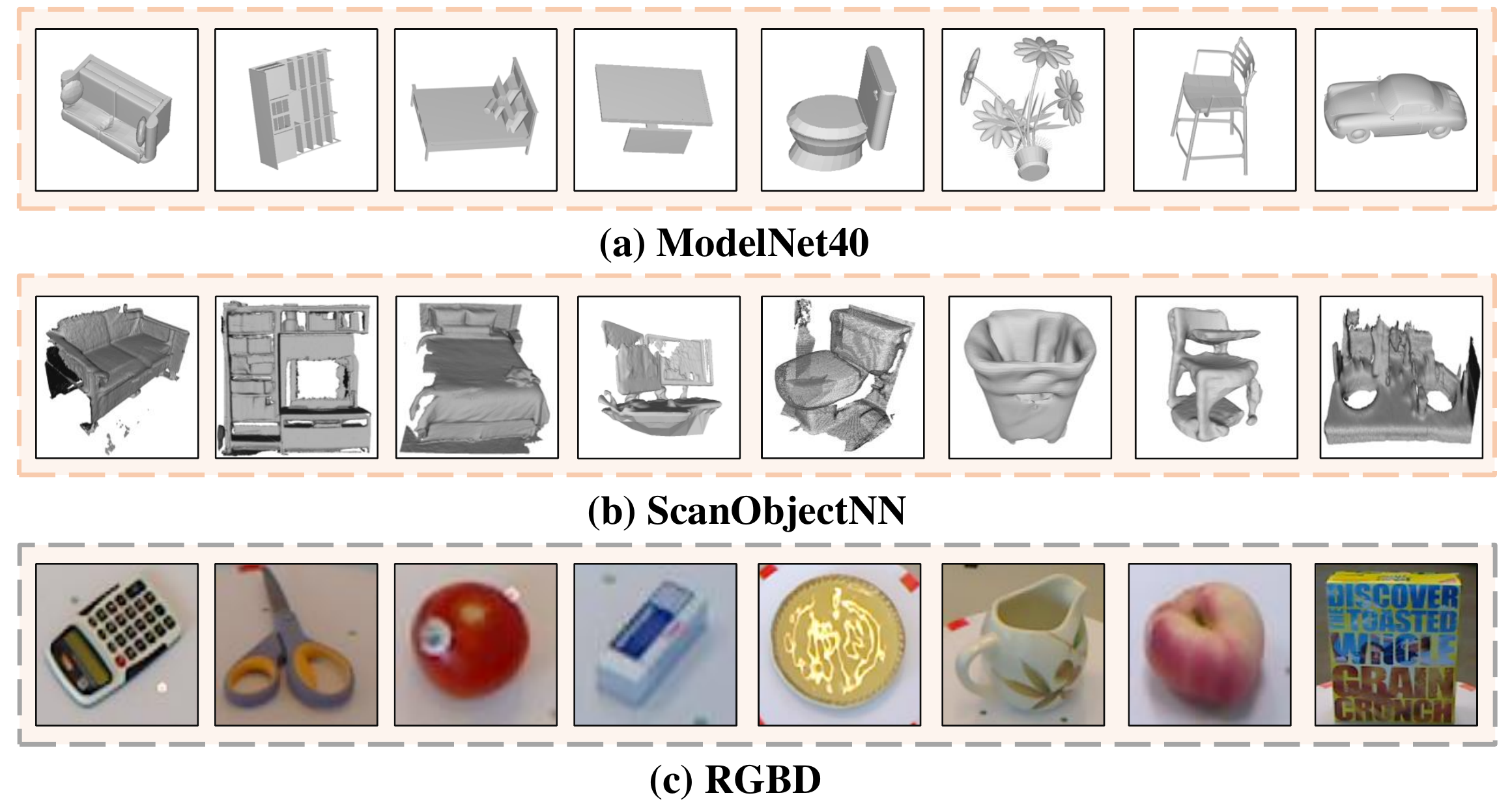}
\caption{Examples of data in three different datasets. }
\label{fig:dataset} 
\end{figure}

\section{Experiments}
We evaluate the performance of our method on multiple datasets including ModelNet40~\cite{modelnet}, ScanObjectNN~\cite{scanobjectnn} and RGBD~\cite{scanobjectnn}, examples of data in these datasets are shown in Fig.~\ref{fig:dataset}. For each dataset, we conduct experiments under the arbitrary view setting and the fixed viewpoint setting, where the 3D objects are either aligned or rotated. To keep the comparisons fair, we re-implement MVCNN~\cite{mvcnn} and GVCNN~\cite{gvcnn} denoted as MVCNN-M and GVCNN-M, and they utilize the exact same backbone network and training settings as ours. 

\subsection{Data Preparation}
\noindent{\textbf{ModelNet40 and ScanObjectNN}}. For the arbitrary view setting on ModelNet40~\cite{modelnet} and ScanobjectNN~\cite{scanobjectnn}, we generate the projected views with the following steps: 
(i) Randomly choose 6 to 20 points from a spherical surface as camera locations. (ii) Project the object from the chosen viewpoints to obtain the 2D views (cameras are assumed to point to the centroid of the object). For the fixed viewpoint setting with object rotation, we obtain the 2D views by first  
rotating the object around X-axis by a random angle between 0 and 180 degrees and then projecting the object from 20 fixed viewpoints that constitute a dodecahedron similar to~\cite{rotationnet,view-gcn}. As for the aligned setting, we follow the setup used in other work like ~\cite{rotationnet,view-gcn}. We compare our performance with the state-of-the-art methods applicable to the specific setting. Note that since ScanObjectNN provides 3D models in the form of point clouds, we first reconstruct them into meshes with Poisson Surface Reconstruction~\cite{poisson}.

\noindent{\textbf{RGBD dataset}}. The RGBD dataset~\cite{rgbd} contains real-world pictures of objects from a large number of viewpoints, without providing 3D scans of these objects. Thus we simulate the arbitrary view settings by randomly sampling 4 to 12 images from each object instance. We also perform 10-fold cross validation on this dataset. In each round, we randomly leave one instance from each class out for testing while the rest are used in training. 

\subsection{Experiments on ModelNet40}
This dataset consists of 12,311 3D shapes from 40 categories, with 9,483 training models and 2,468 test models for shape classification. It is the most widely adopted benchmark for 3D shape classification. Various methods reported results on this dataset using different shape representations including voxels, point clouds and multi-view images. 

The experimental results on ModelNet40~\cite{modelnet} are shown in Tab.~\ref{table:modelnet}. Among the previous methods, view-GCN~\cite{view-gcn} and RotationNet~\cite{rotationnet} are two powerful methods and produce state-of-the-art results when the objects are aligned. However, their classification accuracies drop dramatically by more than 9.3\% under the rotated object setting, in which the projected 2D images are not well aligned. Our method outperforms them by 3.97\% per class and 5.22\% per instance accuracy, showing that our method can obtain more robust representations from perturbed objects.

For the arbitrary view setting, we can see that our proposed method achieves notably better accuracy than the compared methods with the margin of 3.34\% per instance and 3.03\% per class accuracy. Note that RotationNet~\cite{rotationnet} and view-GCN~\cite{view-gcn} are not applicable to the arbitrary view setting because RotationNet~\cite{rotationnet} assumes pre-defined viewpoints while view-GCN~\cite{view-gcn} requires given fixed viewpoint positions to construct view-graph in training and testing.

%

\subsection{Experiments on ScanObjectNN}
ScanObjectNN~\cite{scanobjectnn} is a recently proposed real-world 3D object classification dataset with scanned indoor scene data. It contains around 15000 objects that are categorized into 15 categories with 2902 unique object instances. ScanObjectNN offers more practical challenges including background occurrence, object partiality, and different deformation variants.
The results on ScanObjectNN are shown in Tab.~\ref{table:scanobject}. Under the arbitrary view setting, our approach significantly outperforms MVCNN-M and GVCNN-M by more than 6.01\% and 9.23\% on per-instance and per-class accuracy respectively. As for the fixed viewpoint setting, while our approach performs similarly with the current state-of-the-art on aligned objects, it achieves better results on rotated objects, improving per-instance and per-class accuracy by 2.09\% and 2.71\%. 
\begin{table}[]
\centering
\caption{Shape classification accuracy (in\%) on RGBD.}
\small\setlength{\tabcolsep}{3mm}
{\begin{tabular}{l c c c}
\toprule
\hline
Method      &Setting     & $\#$View                   & Per Ins. Acc. \\\hline
MDSICNN~\cite{MDSICNN}     &  \multirow{5}{*}{Fixed}      & $\geq$  120 & 89.6 \%        \\
CFK\cite{CFK}        &        & $\geq$  120  & 86.8   \%      \\
MMDCNN~\cite{MMDCNN}       &      &  $\geq$  120   & 86.8 \%        \\
RotationNet~\cite{rotationnet} &  & 12                  & 89.3\%     \\
View-GCN~\cite{view-gcn}     &        &            12           & \textbf{94.3}\%         \\\hline
MVCNN-M       &    \multirow{3}{*}{Arbitrary}     & \multirow{3}{*}{4-12}                     &   89.0\%       \\
GVCNN-M       &         &                     & 89.8 \%            \\
Ours         &        &                      &  \textbf{91.8}\%         \\\hline
\bottomrule
\end{tabular}}\label{table:RGBD}
\end{table}

\subsection{Experiments on RGBD Dataset}
To further evaluate our method for recognizing real captured multi-view images, we conduct experiments on multi-view shape recognition with images from the RGBD dataset~\cite{rgbd}. We randomly select images captured by camera with different elevation angles and the view number is varying from 4 to 12. As shown in Tab.~\ref{table:RGBD}, for the arbitrary view setting, our method outperforms MVCNN-M and GVCNN-M by 2\%, which shows that our method is  capable of dealing with real multi-view images captured from arbitrary views. Our method in arbitrary view setting (4 to 12 views) also exceeds the results of RotationNet in the setting of fixed 12 views.

\subsection{Ablation Study}\label{sec:ablation}
In this section, we take a closer look at the effects of key components of our network. Experiments are conducted on ModelNet40 under the arbitrary view setting.

\noindent{\textbf{Effects of image-level feature encoder.}} We examine the effects of the Image-level Feature Encoder (ILFE) defined in Sect.~\ref{sec:ILF-encoder}. We compare it with a baseline network that extracts features with a CNN and aggregates them with Max-pooling, the same structure as MVCNN~\cite{mvcnn}. To evaluate the effects of the ILFE, We use the ILFE instead of the CNN to extract the view features. As shown in Tab.~\ref{table:ablation_modules_old}, the ILFE brings 1.92\% and 2.92\% improvement on per-instance and per-class accuracies over the baseline. This proves the effectiveness of the Transformer that explores the relationship among arbitrary views. 


\noindent{\textbf{Empirical analysis of canonical view representation.}} We evaluate the effects of the Canonical View Representation~(CVR) module defined in Sect.~\ref{sec:CR-decoder}, and our choice of using optimal transport to obtain canonical view features. As shown in Tab.~\ref{table:CVR}, if we remove the CVR module completely, the per-instance and per-class accuracies drop by 1.87\% and 2.00\% respectively. Aside from optimal transport, Transformer decoder~\cite{attention} is a popular network structure that can align an arbitrary number of input features into fixed sized features. While structurally identical to a Transformer encoder, a Transformer decoder takes learnable reference view features $Z$ as query and image-level features $F^0$ as key and value. We substitute the CVR module with a Transformer decoder while the rest of the network is kept unchanged. Surprisingly, the results further drop by 2.04\% and 1.75\%. This shows that optimal transport is a superior way to align features and brings a significant performance boost to our network.

\begin{figure} 
\centering 
\includegraphics[width=0.5\textwidth]{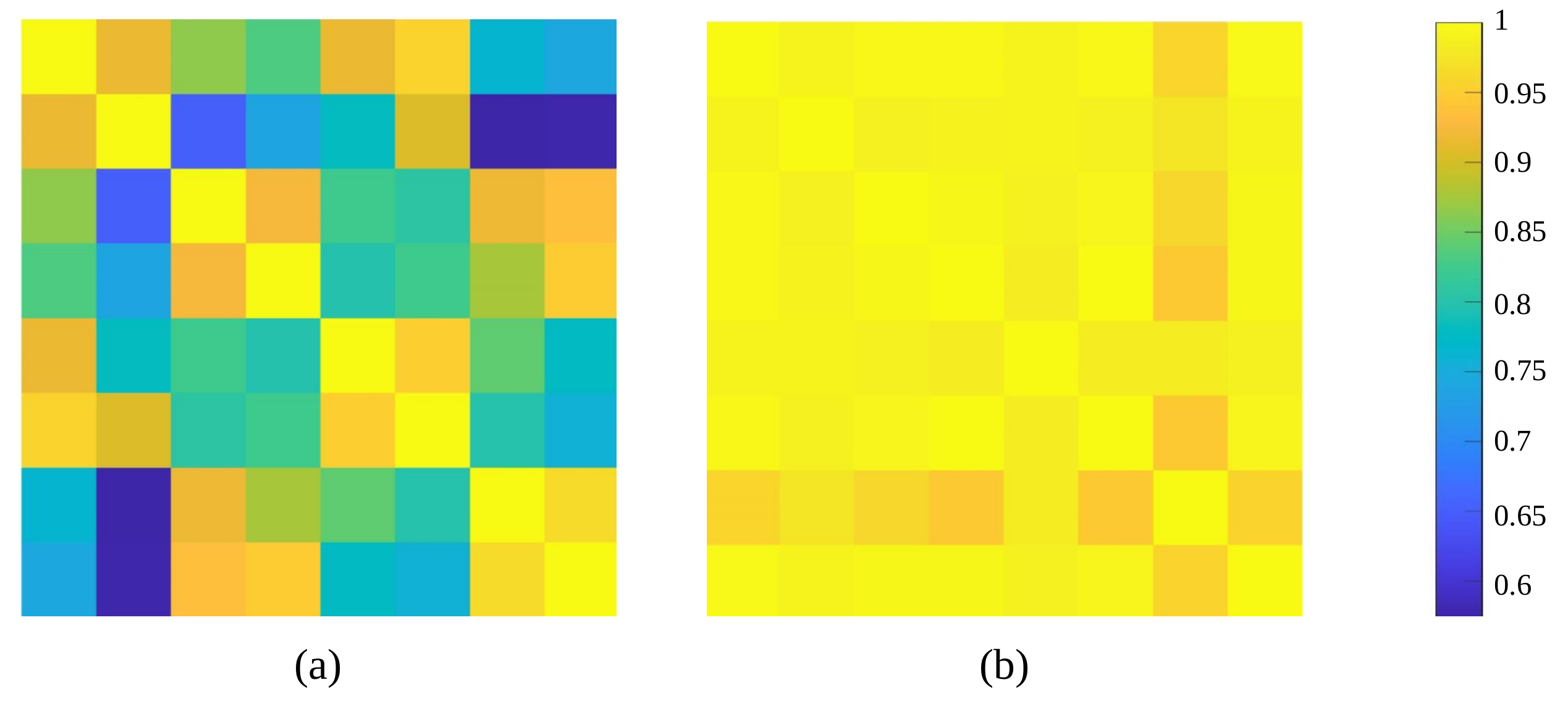}
\caption{Visualization of canonical view features $F^c$ with~(a) and without (b) the canonical view feature separation constraint. Each element of the matrix is the cosine similarity of paired canonical view features.}  
\label{fig:spatial_loss} 
\end{figure}


\noindent{\textbf{Effects of canonical view feature separation constraint.}} We now study empirically how the Canonical View Feature Separation Loss (CVFSL) affects the performance. As shown in Tab.~\ref{table:ablation_modules_old}, with CVFSL, our method achieves notably better results with 1.59\% and 1.73\% improvements on per-instance and per-class accuracies. 
Moreover, we can observe in Fig.~\ref{fig:spatial_loss}~(b) that without CVFSL, the resulting features have a flat similarity matrix. This means that canonical view features are not properly distinguishable in the feature space, resulting in a non-informative representation of the 3D shape. With CVFSL, the features are much more diverse as shown in (a), which can explain the larger performance gain with the CVFSL enabled. We also compare it with a cosine similarity loss that simply forces canonical view features to be different. The results drop by 1.06\% and 1.21\% in two accuracies. Therefore, we conclude that the canonical view feature separation constraint is vital in obtaining an informative canonical view representation and a robust final feature representation of the 3D shape.


\noindent{\textbf{Selection of the number of reference view features.}} In this paper, we have introduced the reference view features to which the features from arbitrary views are aligned. Here we evaluate the effect  of the number ($M$) of reference view features. As shown in Tab.~\ref{table:choiceofM}, different choices of $M$ results in notable differences in performance on ModelNet40 with arbitrary views. Specifically, $M=8$ results in the best performance, followed by $M=16$, and $M=4$ at last. We can infer that on the arbitrary-view setting of ModelNet40, $M=8$ is preferable. Note that this result may vary on different datasets and settings with different distributions of viewpoints. 

\noindent{\textbf{Effects of the canonical view aggregator.}} Now we examine the effects of the Canonical View Aggregator (CVA) module defined in Sect.~\ref{sec:CF-aggregator}. We remove the CVA from our network, instead we directly perform view pooling on the output features from CVR. Comparing the results shown in Tab.~\ref{table:ablation_modules_old}, we find that removing the CVA leads to a performance drop of 1.16\% and 1.97\% on per-instance and per-class accuracy. Therefore, it is shown that the CVA module, which makes use of the aligned spatial encoding and further models the relationship among features in canonical representation, is crucial to the performance of our network.


\begin{table}[]
\caption{Comparison of optimal transport with Transformer decoder in CVR.}
\centering
\label{table:CVR}
 \small\setlength{\tabcolsep}{1.5mm}
{\begin{tabular}{ccc}
\toprule
\hline
                  & Per Class Acc. & Per Ins. Acc. \\ \hline
w/o CVR                      & 82.01\%         & 85.04\%       \\ \hline            
Transformer decoder           & 79.97\%         & 83.29\%       \\ \hline
Optimal transport (ours)       & \textbf{84.01}\%         & \textbf{86.91}\%       \\ \hline
\bottomrule
\end{tabular}}
\end{table}

\begin{table}[]
\centering
\caption{Ablation study on each module of our network.}
\label{table:ablation_modules_old}
\small\setlength{\tabcolsep}{1mm}
{\begin{tabular}{cccccc}
\toprule
\hline
ILFE                       & CVR            & CVA                      & CVFSL                       & Per Class Acc.  & Per Ins. Acc.    \\ \hline
                          &                           &                           &                           & 78.86\%          & 83.20\%          \\ \hline
\checkmark &                           &                           &                           & 81.78\%          & 85.12\%          \\ \hline
\checkmark &               \checkmark            &  &      \checkmark                     & 82.04\%          & 85.75\%          \\ \hline
\checkmark & \checkmark & \checkmark &                           & 82.28\%          & 85.32\%          \\ \hline
\checkmark & \checkmark & \checkmark & \checkmark & \textbf{84.01\%} & \textbf{86.91\%} \\ \hline
\bottomrule
\end{tabular}}
\end{table}


\begin{table}[]
\centering
 \caption{Results by choosing different number of reference view features.} 
 \label{table:choiceofM}
 \small\setlength{\tabcolsep}{5mm}
 {\begin{tabular}{ccc}
\toprule
\hline
            & Per Class Acc. & Per Ins. Acc. \\ \hline
M = 4  & 82.41\%         & 85.69\%       \\ \hline
M = 8  & \textbf{84.01\%}         & \textbf{86.91\%}       \\ \hline
M = 16 & 82.71\%         & 86.22\%       \\ \hline
\bottomrule
\end{tabular}}
\end{table}

\section{Conclusion and discussion}
In this work, we propose a novel canonical view representation to tackle the challenge of 3D shape recognition with arbitrary views. We incorporate optimal transport with the canonical view feature separation constraint to transform the features of arbitrary views into an aligned canonical view representation, enabling us to aggregate and derive a rich and robust feature representation for the 3D shape. The experimental results prove the effectiveness of our method. As discussed in Sect.~\ref{sec:CR} and Sect.~\ref{sec:net}, the learned reference view features in $Z$ set up a common reference for aligning arbitrary views to a fixed number of learnable reference views. This approach can potentially be applied to other multi-view vision tasks, such as view synthesis, view-based 3D reconstruction or generation, which we may investigate in our future work.  

{\textbf{Acknowledgment}} This work was supported by NSFC with grant numbers of 11971373, U20B2075, 11690011, U1811461, 12026605, 12090021, 61721002, and National Key R$\&$D Program 2018AAA0102201.
{\small
\bibliographystyle{ieee_fullname}
\bibliography{egbib}
}

\end{document}


\title{Learning Canonical View Representation for 3D Shape Recognition with Arbitrary Views\\
{\large Supplementary Material}}

\maketitle
\ificcvfinal\thispagestyle{empty}\fi

In this supplementary material, we provide additional ablation study and the visualization for our approach. 

\section*{A.~Additional Ablation Study}
To further evaluate the performance impact of different components in our network, we report additional results on the selections of hyperparameters and architectures. We conduct all these experiments on ModelNet40~\cite{modelnet} under the arbitrary-view setting. 
\section*{A.1.~Backbone network}
We first examine the performance of our method with different CNN backbone networks: AlexNet~\cite{alexnet}, ResNet-18~\cite{resnet}, ResNet-50~\cite{resnet} and ResNet-101~\cite{resnet}. As shown in Tab.~\ref{table:backbone}, a more efficient backbone network produces better performance, as variants of the ResNet architecture outperform AlexNet significantly. However, the performance margin among the ResNet backbones are much less noticeable, with the deeper ResNet-101 achieves less than 1\% gain in accuracy over ResNet-18. We choose the ResNet-18 network for our implementation since it performs reasonably well while being less computationally expensive.

\section*{A.2.~Obtaining the global representation}
As mentioned in Sect.~{4.3}, Global Average Pooling (GAP) is performed on the outputs of the Transformer Encoder in canonical view aggregator to obtain a global representation of the 3D shape. Here we compare the GAP to other methods in producing the global representation, including Global Max Pooling (GMP) and concatenating the features directly. As shown in Tab.~\ref{table:pooling}, we can see that GAP performs noticeably better than GMP, while marginally outperforming the direct concatenation of features. One possible explanation for GMP's lower performance is that the gradients are only back-propagated to the maximum elements. For our particular network design, this could potentially be harmful for learning diverse and robust canonical view features. 

 
 \begin{table}[]
\centering
 \caption{Results with different CNN backbone networks.} 
 \label{table:backbone}
 \small\setlength{\tabcolsep}{5mm} 
 {\begin{tabular}{lcc}
\toprule
\hline
Backbone   & Per Class Acc. & Per Ins. Acc. \\\hline
AlexNet~\cite{alexnet}    & 75.94\%        & 78.88\%      \\\hline
ResNet-18~\cite{resnet}   & 84.01\%        & 86.91\%      \\\hline
ResNet-50~\cite{resnet}  & 83.64\%        & 87.18\%      \\\hline
ResNet-101~\cite{resnet} & \textbf{84.34}\%        & \textbf{87.77}\%     \\\hline
\bottomrule
\end{tabular}}
\end{table}
 
\begin{table}[]
\centering
 \caption{Comparing methods for obtaining global representation.} 
 \label{table:pooling}
 \small\setlength{\tabcolsep}{5.7mm}
 {\begin{tabular}{l c c}
\toprule
\hline
           & Per Class Acc. & Per Ins. Acc. \\\hline
Concat.     & 83.76\%        & 86.42\%      \\\hline
GMP & 82.77\%        & 85.41\%      \\\hline
GAP & \textbf{84.01}\%        & \textbf{86.91}\%     \\\hline

\bottomrule
\end{tabular}}
\end{table}

\begin{table}[]
\centering
 \caption{Impact of the weighting factor $\lambda$ for the Spatial-Awareness Constraint Loss.} 
 \label{table:lambda}
 \small\setlength{\tabcolsep}{5.7mm}
 {\begin{tabular}{l c c}
\toprule
\hline
          & Per Class Acc. & Per Ins. Acc. \\\hline
$\lambda = 1.0$ & 82.97\%        & 85.12\%      \\\hline
$\lambda = 0.5$ & 81.06\%        & 84.02\%      \\\hline
$\lambda = 0.1$ & \textbf{84.01}\%        & \textbf{86.91}\%     \\\hline
\bottomrule
\end{tabular}}
\end{table}
\begin{table}[]
\centering
 \caption{Comparison of learnable spatial embeddings with fixed sinusoidal positional embeddings.} 
 \label{table:fixed}
 \small\setlength{\tabcolsep}{5.7mm}
 {\begin{tabular}{l c c}
\toprule
\hline
          & Per Class Acc. & Per Ins. Acc. \\\hline
Fixed & 82.14\%        & 85.41\%      \\\hline
Learned & \textbf{84.01}\%        & \textbf{86.91}\%     \\\hline
\bottomrule
\end{tabular}}
\end{table}
\begin{figure*}[!htbp]
\centering
\subfigure[MVCNN-M]{
\includegraphics[width=1\columnwidth]{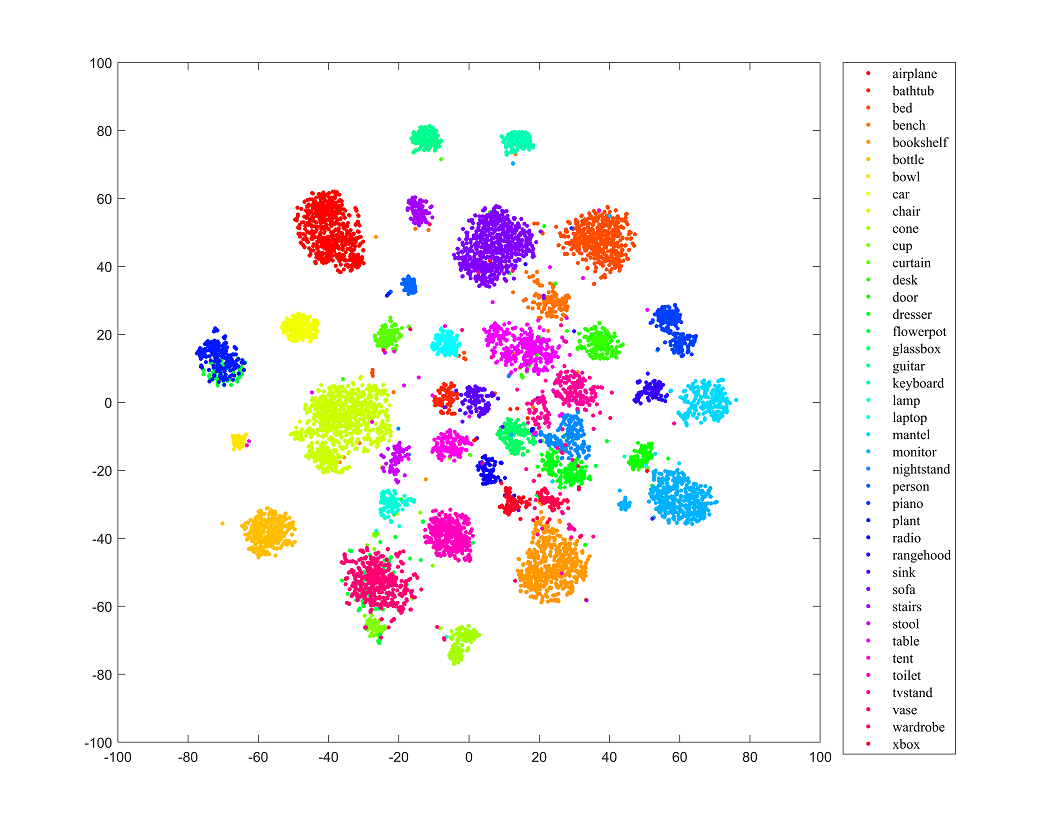}}
\subfigure[Ours]{
\includegraphics[width=1\columnwidth]{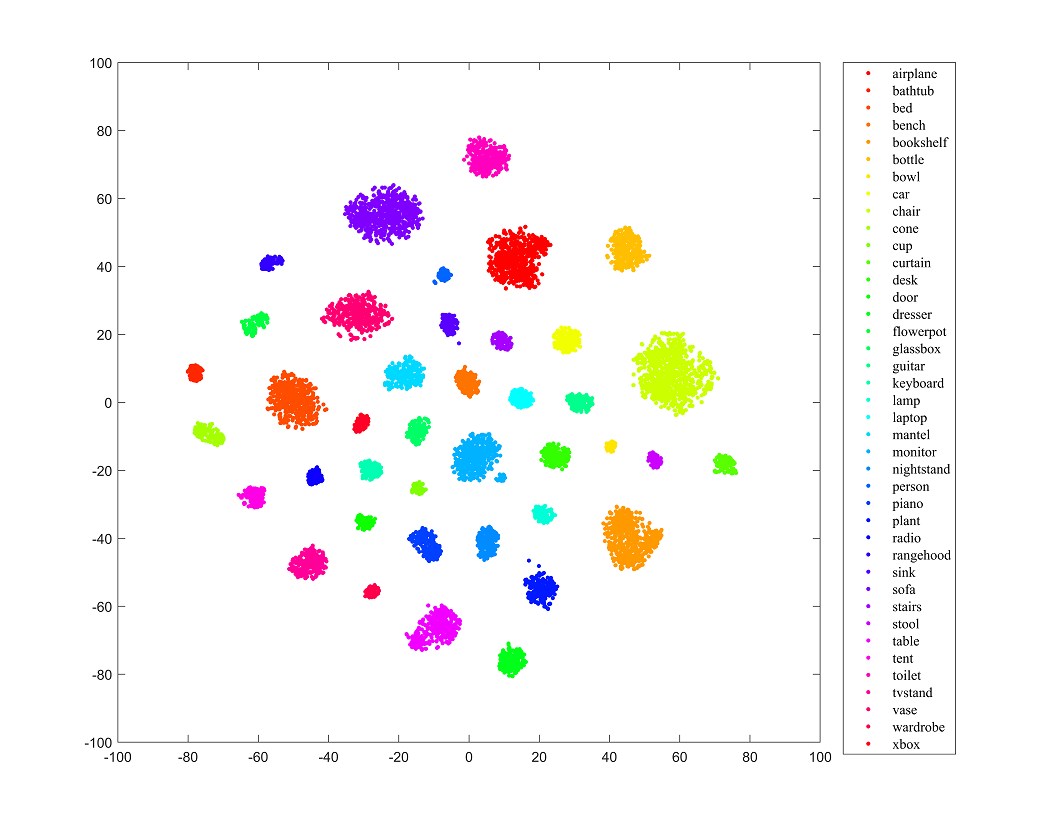}}
\caption{Visualization of shape features learned by MVCNN-M~(a) and our method~(b) via t-SNE on ModelNet40 train set.}
\label{fig:tsne_train} 
\end{figure*}
\begin{figure*}[!htbp]  
\centering
\subfigure[MVCNN-M]{
\includegraphics[width=1\columnwidth]{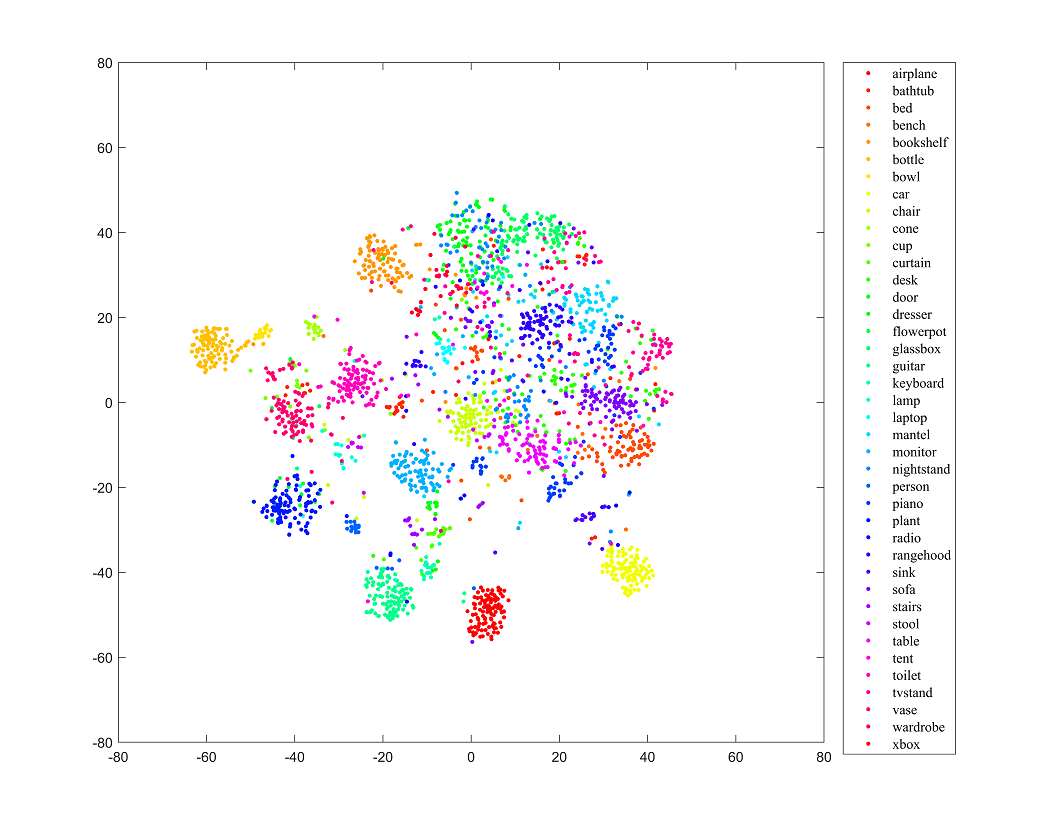}}
\subfigure[Ours]{
\includegraphics[width=1\columnwidth]{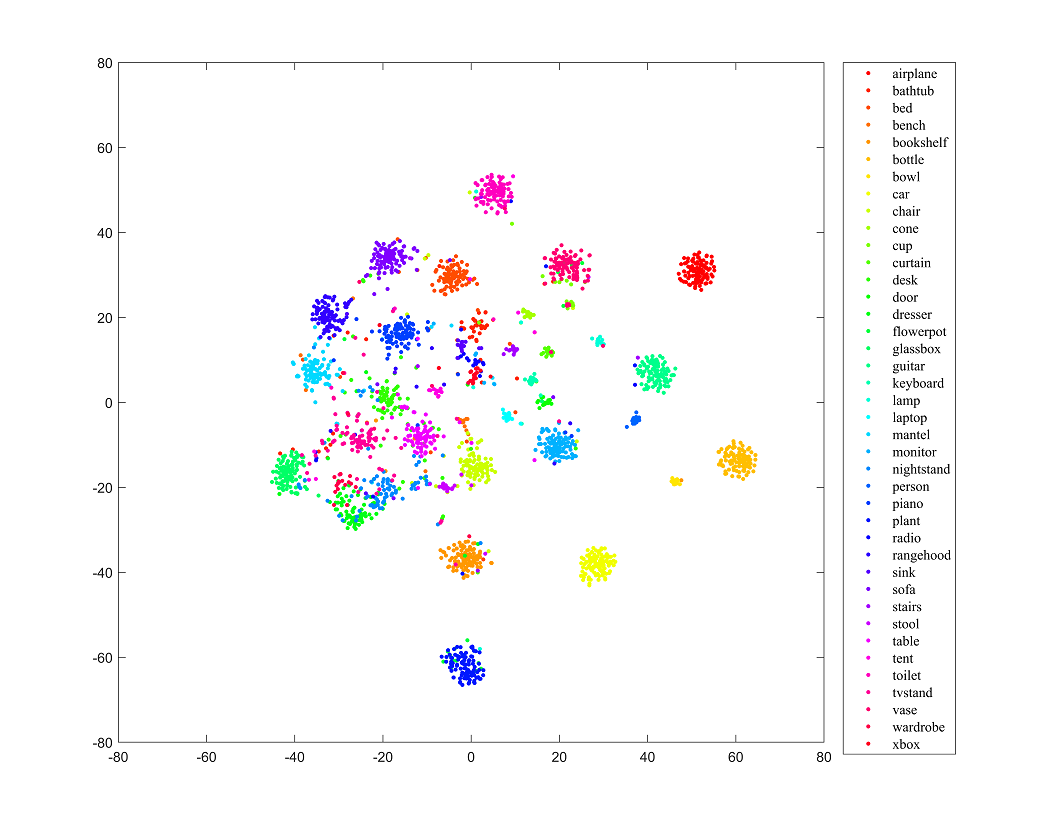}}
\caption{Visualization of shape features learned by MVCNN-M~(a) and our method~(b) via t-SNE on ModelNet40 test set.}
\label{fig:tsne_test} 
\end{figure*}
\section*{A.3.~Loss coefficient $\lambda$}
As defined in Eq.~(11), the overall loss of our network consists of the classification loss $\text{L}_{cls}$ and the Spatial-Awareness Constraint Loss (SACL) $\text{L}_{spatial}$, where the coefficient $\lambda$ controls the weighting factor between the two loss functions. We conduct experiments to examine how $\lambda$ can affect the performance. As seen in Tab.~\ref{table:lambda}, increasing $\lambda$ from 0.1 to 0.5 and 1.0 lowers the classification accuracy. This shows that a good balance between the classification loss and the SACL is important for maximizing the performance. We set $\lambda=0.1$ for our implementation in all experiments.


\section*{A.4.~Positional embedding}
Positional embedding is crucial in Transformer-based architectures to capture sequential information of the inputs. Vaswani et al.~\cite{attention} originally adopts fixed sinusoidal positional embeddings to represent positions, where the $t$-th input's sinusoidal positional embedding is defined as
\begin{equation}\label{eq:fixed}
    \begin{aligned}
    PE_{(t,2i)} = \sin{(t/100000^{2i/d})} \\
    \end{aligned}
\end{equation}where $d$ is the feature dimension and $i=1,2,...,d$.

As mentioned in Sect.~4.3, our approach uses learnable spatial embeddings $F^{se}=\Psi(F^s)$ to encode positional information, where $F^s$ is the spatial representation inferred from the canonical view features $F^c$ by a two-layer MLP $\Psi$ and is constrained by Spatial-Aware Constraint Loss (SACL). To compare the performance impacts of fixed and learned embeddings, we substitute the learned spatial embedding $F^{se}$ with fixed sinusoidal positional embeddings. As shown in Tab.~\ref{table:fixed}, the classification results drop by 1.87\% and 1.50\% in two accuracies, which demonstrates the effectiveness of learnable spatial embeddings.

\section*{B.~Visualization}
In Fig.~\ref{fig:tsne_train} and Fig.~\ref{fig:tsne_test}, we visualize the features learned by MVCNN-M~\cite{mvcnn} and our method on both the train set and the test set of ModelNet40 under the arbitrary-view setting. We perform t-SNE~\cite{tsne} on features of object instances from all classes to visualize the feature discriminability on a macro level. According to Fig.~\ref{fig:tsne_train} and Fig.~\ref{fig:tsne_test}, in both the train set and the test set, features from our method display much better clustered distributions under t-SNE than those produced by MVCNN-M~\cite{mvcnn}. Specifically, we can observe both lower intra-class variance and higher inter-class variance in the results of our method compared to MVCNN-M~\cite{mvcnn}, which reflects better overall shape classification performance on ModelNet40 with arbitrary view. 

{\small
\bibliographystyle{ieee_fullname}
\bibliography{egbib}
}